\documentclass{article}
\PassOptionsToPackage{numbers,sort&compress}{natbib}
\usepackage[preprint]{neurips_2026}

\usepackage[utf8]{inputenc} 
\usepackage[T1]{fontenc}    
\usepackage[breaklinks=true,bookmarks=true,colorlinks,pagebackref=true]{hyperref}
\usepackage{url}            
\usepackage{booktabs}       
\usepackage{amsfonts}       
\usepackage{nicefrac}       
\usepackage{microtype}      
\usepackage{enumitem}
\usepackage{amsmath}
\usepackage{multirow}
\usepackage[table]{xcolor}

\usepackage{tabularx}
\usepackage{tikz}
\usetikzlibrary{positioning,arrows.meta,fit,calc,shadows.blur}
\usepackage{subcaption}
\usepackage{float}
\usepackage{pifont}       
\newcommand{\cmark}{\ding{51}}
\newcommand{\xmark}{\ding{55}}
\usepackage{threeparttable} 
\usepackage{graphicx}      
\usepackage{titlesec}
\titlespacing\section{0pt}{0pt plus 0pt minus 0pt}{0pt plus 0pt minus 0pt}
\titlespacing\subsection{0pt}{0pt plus 0pt minus 0pt}{0pt plus 0pt minus 0pt}
\titlespacing\subsubsection{0pt}{0pt plus 0pt minus 0pt}{0pt plus 0pt minus 0pt}
\usepackage{bbding}

\title{Bench2Drive-Robust: Benchmarking Closed-Loop Autonomous Driving under Deployment Perturbations
}

\author{%
 Zhiyuan Zhang$^{3*}$ \And  Zhenghao Jin$^{1,2}$\thanks{This work was partly done when Zhenghao Jin interns at  Great Wall Motor.} \And  Yanlun Peng$^{2}$ \And  Xianda Guo$^{4}$ \AND  Haoran Liu$^{1}$ \And  Shaofeng Zhang$^{5}$  \And Xingjun Ma$^{1}$  \And  Zuxuan Wu$^{1}$  \AND  Junchi Yan$^{3}$\textsuperscript{\Envelope}  \quad\quad\quad  Xiaosong Jia$^{1}$\textsuperscript{\Envelope} \quad\quad Yu-Gang Jiang$^{1}$ 
 \\ \\
$^{1}$ Institute of Trustworthy Embodied AI (TEAI), Fudan University \\
$^{2}$ Great Wall Motor \\
$^{3}$ Sch. of Computer Science \& Sch. of Artificial Intelligence, Shanghai Jiao Tong University\\
$^{4}$ School of Computer Science, Wuhan University \\
$^{5}$ University of Science and Technology of China \\
* Equal Contributions \quad\quad
\textsuperscript{\Envelope} Correspondence Author
\\
\normalsize{
    \url{https://github.com/Thinklab-SJTU/Bench2Drive-Robust}
    }
}

\begin{document}

\maketitle

\begin{abstract}
Robustness is a critical requirement for deploying autonomous driving systems in the real world. Existing robustness benchmarks for autonomous driving have made important progress in studying the effects of image-level corruptions, such as adverse weather or camera degradation, on perception modules and open-loop planning outputs. 
However, deployment can also involve system-level imperfections, such as inference latency and ego-state estimation errors, which remain less studied in closed-loop E2E-AD evaluation.
These imperfections can accumulate through the feedback loop and destabilize control.
In this work, we present \textbf{Bench2Drive-Robust}, to our knowledge the first device-centric robustness benchmark for closed-loop end-to-end autonomous driving under realistic deployment perturbations. We systematically evaluate deployment-oriented perturbations arising from three major sources: camera-stream failures (frame drop, partial observation), ego-state estimation errors (GPS noise, and speed or odometry errors), and compute-induced control delay (model inference delay). 
We evaluate representative end-to-end driving methods and analyze their robustness under different perturbation severities. Our results show that these deployment-related perturbations can substantially degrade closed-loop driving performance, revealing robustness challenges that are not fully captured by conventional image-level corruption evaluations. By establishing a closed-loop evaluation protocol and demonstrating the substantial impact of these deployment-oriented perturbations, Bench2Drive-Robust defines practical robustness problems for end-to-end autonomous driving and encourages further research on deployment-aware robust driving systems.

\end{abstract}

\begin{figure*}[!t]
    \centering
    \includegraphics[width=\linewidth]{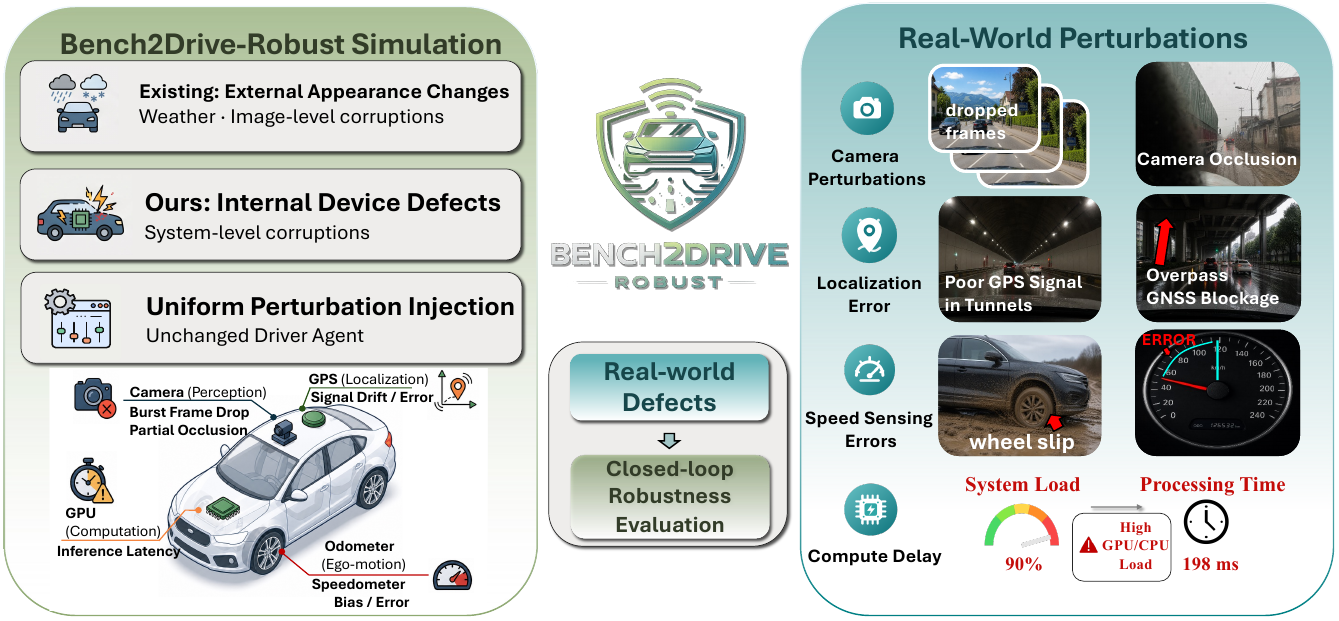}
    \caption{
        \textbf{Overview of Bench2Drive-Robust.}
        We evaluate three categories of deployment-side failures for E2E-AD—camera-stream failures, ego-state estimation errors, and compute-induced control delay—under closed-loop driving. This differs from existing image- or perception-centric robustness benchmarks that mainly target external appearance changes.
    }
    \label{fig:teaser}
    \vspace{-10pt}
\end{figure*}

\section{Introduction}
\label{sec:introduction}

Autonomous driving systems are expected to operate under diverse and imperfect real-world conditions. 
Recent end-to-end autonomous driving (E2E-AD) methods have achieved strong performance by mapping sensor observations directly to planning trajectories or control actions~\citep{wu2022tcp,hu2023uniad,jiang2023vad,chitta2022transfuser,renz2025simlingo,jia2025drivetransformer}. However, existing closed-loop evaluations typically assess driving capability under diverse scenarios and weather conditions while assuming a reliable simulator--policy interface.

Robustness under deployment conditions is an important step toward practical E2E-AD systems. 
As illustrated in Fig.~\ref{fig:teaser}, real vehicles may encounter camera stream failures, localization errors, speed sensing errors, and compute-control delays caused by onboard hardware, sensing, and software-stack imperfections. 
Studying such perturbations can provide feedback for onboard system design and encourage E2E-AD algorithms that are more tolerant to realistic software- and device-side variations.

However, these factors affect not only the visual content observed by the model, but also the timing, consistency, and reliability of the perception--control loop.
Existing robustness benchmarks have made important progress in studying appearance- and perception-level corruptions, including adverse weather, sensor noise, depth degradation, LiDAR corruptions, and BEV perception failures~\citep{dong2023benchmarking,kong2023robo3d,kong2023robodepth,xie2025robobev,mapbench2024}.
While these evaluations provide valuable diagnostic signals, they often focus on perception modules or open-loop planning outputs, and therefore may not fully capture how perturbations propagate through vehicle motion and future observations in closed-loop driving~\citep{jiang2024robuste2e,robodrivebench2025,zhai2023admlp,jia2024bench2drive}.
This motivates a complementary closed-loop benchmark that evaluates how E2E-AD models respond to deployment-side system perturbations during interactive driving.

In this work, we present \textbf{Bench2Drive-Robust}, a device-centric closed-loop robustness benchmark for E2E-AD built on Bench2Drive~\citep{jia2024bench2drive}. 
Our benchmark translates real-world system imperfections into parameterized simulation perturbations covering three deployment-relevant sources: camera-stream failures, including frame drop and partial observation; ego-state estimation errors, including GPS noise and speed errors; and compute-induced control delay, including inference latency.
We evaluate representative E2E-AD models, including TCP~\citep{wu2022tcp}, UniAD~\citep{hu2023uniad}, VAD~\citep{jiang2023vad}, and SimLingo~\citep{renz2025simlingo}, under increasing perturbation severities. 

In summary, Bench2Drive-Robust features:
\vspace{5pt}
\begin{itemize}[leftmargin=10pt, topsep=0pt, itemsep=1pt, partopsep=1pt, parsep=1pt]
    \item A closed-loop robustness benchmark for E2E-AD under deployment-side perturbations beyond standard image-level corruptions.
    
    \item A device-centric perturbation taxonomy covering camera-stream failures, ego-state estimation errors, and compute-induced control delay.

    \item A modular perturbation injection framework with parameterized severity settings, enabling structured stress tests across camera-stream, ego-state, and compute-control failures.
    
    \item A systematic evaluation of representative E2E-AD models under multiple perturbation severities, showing distinct closed-loop robustness profiles.
\end{itemize}

\section{Related Work}

\label{sec:related_work}

\subsection{End-to-End Autonomous Driving}

\label{subsec:rw_e2e_ad}
End-to-end autonomous driving (E2E-AD) aims to learn driving policies that map onboard observations directly to planning trajectories or low-level control commands.
Early neural driving systems studied direct sensorimotor control, conditional imitation learning, privileged distillation, and compact closed-loop policies~\citep{bojarski2016end,codevilla2018cilrs,chen2020learning,prakash2021multi,chitta2022transfuser,wu2022tcp,renz2022plant,shao2023interfuser}.
Recent methods further scale E2E-AD through unified perception--prediction--planning architectures, vectorized and sparse scene representations, BEV/occupancy modeling, and transformer-based policies~\citep{hu2023uniad,jiang2023vad,sun2024sparsedrive,zhang2024sparsead,jia2025drivetransformer,philion2020lift,li2022bevformer,huang2021bevdet,li2023bevdepth,liu2023bevfusion,wang2022detr3d,liu2022petr,wang2023streampetr,zhang2023occformer,wei2023surroundocc,weng2024paradrive}.
Planner-centric advances also explore generative trajectory modeling, flow matching, reinforcement learning, trajectory selection, and temporal-consistency mechanisms~\citep{liao2025diffusiondrive,xing2025goalflow,liu2025guideflow,li2024hydramdp,song2025momad,yao2025drivesuprim,yang2025raw2drive}.
With the rise of foundation models, language-augmented and vision-language-action driving systems have been developed for instruction following, reasoning, interpretability, and action generation~\citep{shao2024lmdrive,sima2024drivelm,ma2024dolphins,drivegpt4,wen2023dilu,renz2025simlingo,orion2025,zhou2025opendrivevla,zhou2025autovla,yang2025drivemoe,jiang2025vla4ad,zheng2025world4drive,lu2025realad}.
Together, these works have advanced E2E-AD model design and driving capability, while our work focuses on evaluating their robustness under deployment-side closed-loop perturbations.

\begin{table}[t]
\centering
\resizebox{\linewidth}{!}{%
\begin{tabular}{llccc}
\toprule
\textbf{Benchmark} & \textbf{Evaluation Level} & \textbf{Closed-Loop} & \textbf{Deployment-Oriented} \\
\midrule
KITTI-C/nuScenes-C/Waymo-C~\citep{dong2023benchmarking} & 3D Object Detection & \xmark & \xmark \\
Robo3D~\citep{kong2023robo3d}                         & 3D Perception         & \xmark & \xmark \\
RoboDepth~\citep{kong2023robodepth}                   & Depth Estimation      & \xmark & \xmark \\
RoboBEV~\citep{xie2025robobev}                        & BEV Perception        & \xmark & \cmark$^\dagger$ \\
RobustE2E~\citep{jiang2024robuste2e}                  & E2E Driving           & \cmark$^\dagger$ & \xmark \\
RoboDriveBench~\citep{robodrivebench2025}             & VLM E2E Driving       & \xmark & \cmark$^\dagger$ \\
RoboDrive Challenge~\citep{kong2024robodrive}         & Multi-Task Perception & \xmark & \xmark \\
MapBench~\citep{mapbench2024}                         & HD Map Construction   & \xmark & \xmark \\
Fail2Drive~\citep{gerstenecker2026fail2drive}         & E2E Driving           & \cmark & \cmark$^\dagger$ \\
\midrule
\textbf{Bench2Drive-Robust (Ours)}                    & E2E Driving           & \cmark & \cmark \\
\bottomrule
\end{tabular}%
}
\caption{
Comparison with existing AD robustness benchmarks.
$^\dagger$ denotes partial coverage.
}
\label{tab:benchmark_comparison}
\vspace{-4pt}
{\footnotesize
RoboBEV includes frame drop and camera crash but remains perception-level. 
RobustE2E includes limited closed-loop cases. 
RoboDriveBench covers sensor/communication failures but does not target temporal or ego-state perturbations.
Fail2Drive focuses on paired scenario-level distribution shifts.
}
\vspace{-20pt}
\end{table}

\subsection{Datasets, Benchmarks, and Robustness Evaluation}
\label{subsec:rw_benchmarks_robustness}

Large-scale datasets have driven progress in autonomous driving perception, prediction, mapping, and planning~\citep{geiger2012kitti,cordts2016cityscapes,huang2018apolloscape,chang2019argoverse,caesar2020nuscenes,sun2020waymo,yu2020bdd100k,mao2021once,xiao2021pandaset}.
Many logged-data evaluations are open-loop, comparing predictions against recorded expert behavior without affecting future observations, which may not fully reflect interactive driving performance~\citep{zhai2023admlp,dauner2023parting}.
To evaluate planning and control more directly, recent benchmarks introduce closed-loop or simulation-based protocols, including CARLA, nuPlan, NAVSIM, Bench2Drive, Fail2Drive, and DriveArena~\citep{dosovitskiy2017carla,caesar2021nuplan,dauner2024navsim,jia2024bench2drive,gerstenecker2026fail2drive,yang2025drivearena}.
These benchmarks support assessment of route completion, safety, planning quality, generalization, and multi-ability closed-loop behavior for learning-based planners and E2E-AD systems~\citep{huang2023gameformer,weng2024paradrive,li2024hydramdp,bench2drive_speed2026,song2025momad,li2025hydranext,yang2025raw2drive}.

Robustness benchmarks further evaluate autonomous driving systems under distribution shifts, corruptions, and adverse conditions~\citep{hendrycks2018benchmarking,michaelis2019winter}.
Existing protocols mostly focus on perception-side robustness, including image, LiDAR, depth, BEV, 3D perception, and map-construction corruptions~\citep{dong2023benchmarking,kong2023robo3d,kong2023robodepth,xie2025robobev,mapbench2024}.
Recent studies extend robustness evaluation to vision-language driving systems~\citep{jiang2024robuste2e,kong2024robodrive,robodrivebench2025,gerstenecker2026fail2drive}, but still mainly emphasize perception-side corruptions, adversarial or natural corruptions, perception-oriented OoD tracks, and scenario-level distribution shifts~\citep{xie2025robobev,kong2024robodrive,jiang2024robuste2e,gerstenecker2026fail2drive}.
As summarized in Table~\ref{tab:benchmark_comparison}, non-camera interface imperfections such as localization noise, ego-motion sensing errors, and compute-control latency remain underexplored in closed-loop E2E-AD.
This motivates a complementary benchmark that evaluates these perturbations under closed-loop E2E-AD.
\section{Method}
\label{sec:method}
Bench2Drive-Robust is a closed-loop robustness evaluation framework built on Bench2Drive~\citep{jia2024bench2drive}. 
Given a fixed end-to-end driving model, our framework injects controlled perturbations into the closed-loop evaluation pipeline without changing the agent architecture, checkpoint, or decision logic. 
We organize deployment-relevant perturbations into three families: camera-stream failures, ego-state estimation errors, and compute-control delays.
For each perturbation type, we define a severity parameter and apply it consistently across models and routes, enabling model-agnostic comparison under matched closed-loop scenarios.
\begin{figure*}[!t]
    \centering
    \includegraphics[width=\linewidth]{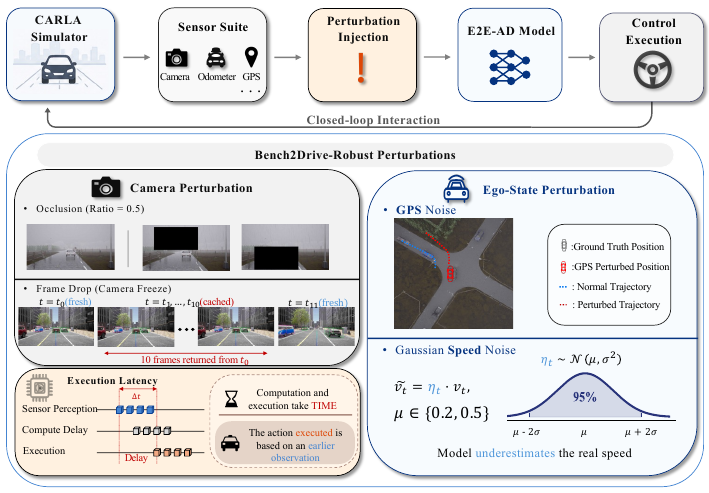}
    \caption{
    \textbf{Overview of Bench2Drive-Robust}. We evaluate end-to-end autonomous driving models in closed-loop simulation by injecting deployment-relevant perturbations into the sensing, ego-state, and action pipelines while keeping the evaluated policy unchanged. The benchmark covers temporal delays, observation integrity perturbations, and ego-state estimation errors, enabling controlled robustness evaluation under consistent driving scenarios.
    }
    \label{fig:method_overview}
    \vspace{-17pt}
\end{figure*}

\subsection{Problem Formulation}
\label{subsec:problem_formulation}

At each simulator timestep $t$, an end-to-end driving policy $\pi$ receives 
visual observations $I_t$ and ego-state signals $s_t$, and outputs a control 
action
\begin{equation}
a_t = \pi(I_t, s_t).
\end{equation}
To evaluate robustness, we perturb the policy inputs before they are consumed 
by the model, yielding perturbed observations $\tilde{I}_t$ and perturbed 
ego-state signals $\tilde{s}_t$. The resulting action is
\begin{equation}
\tilde{a}_t = \pi(\tilde{I}_t, \tilde{s}_t).
\end{equation}

We focus on deployment-oriented imperfections from system constraints and sensor-interface failures, including stale camera streams, noisy ego-state inputs, and delayed control execution. 
We do not add separate weather or lighting corruptions, as such appearance-level variations are already covered by the original Bench2Drive scenarios~\citep{jia2024bench2drive}. 
Our benchmark instead extends Bench2Drive with controlled interface-level perturbations for closed-loop robustness evaluation.

\subsection{Design Goals}
\label{subsec:design_goals}

Bench2Drive-Robust extends Bench2Drive by introducing deployment-oriented robustness as an additional closed-loop evaluation dimension. 
Instead of changing the evaluated E2E-AD models, we perturb the surrounding driving pipeline, including sensor inputs, ego-state estimates, and action timing. 
Each perturbation is controlled by an interpretable severity parameter and applied consistently across models and routes, enabling comparable robustness evaluation under matched scenarios. 
Table~\ref{tab:perturbation_summary} summarizes the perturbation types, severity parameters, and evaluated ranges.

\subsection{Robustness Taxonomy}
\label{subsec:taxonomy}

We organize deployment-oriented perturbations by where they enter the closed-loop driving stack:
\emph{camera-stream perturbations}, \emph{ego-state perturbations}, and \emph{compute-control perturbations}.
This taxonomy separates failures in visual data delivery, vehicle-state sensing, and action execution.
Unlike appearance-level corruptions such as fog, rain, or blur, our perturbations mainly target the interface between the simulator and the driving policy: sensory information may be stale, incomplete, noisy, or delayed even when the external scene remains semantically unchanged.
Table~\ref{tab:perturbation_summary} summarizes the perturbation families and severity settings used in our evaluation.

\begin{table}[t]
\centering
\caption{\textbf{Summary of deployment-oriented perturbations evaluated in Bench2Drive-Robust.}}
\label{tab:perturbation_summary}
\footnotesize
\setlength{\tabcolsep}{5pt}
\renewcommand{\arraystretch}{1.12}
\begin{tabularx}{\linewidth}{l l X X}
\toprule
\textbf{Category} & \textbf{Perturbation} & \textbf{Severity setting} & \textbf{Modeled failure} \\
\midrule

Camera stream
& Burst frame drop
& Burst length $k\in\{1,3\}$ s
& Stream interruption / frozen camera \\

Camera stream
& Partial observation
& Mask ratio $r\in\{0.5,0.8\}$
& Partial camera occlusion \\

\midrule

Ego-state
& GPS noise
& Std. $\sigma_{\mathrm{GPS}}\in\{5,15\}$ m
& Noisy localization input \\

Ego-state
& Speed noise
& $\eta_t\sim\mathcal{N}(\mu,0.2^2)$, $\mu\in\{0.2,0.5\}$
& Speed sensing error \\

\midrule

Compute-control
& Inference latency
& Delay $\tau\in\{100,200,500\}$ ms
& Delayed action execution \\

\bottomrule
\end{tabularx}
\vspace{-6pt}
\end{table}

\subsubsection{Inference Latency}
\label{subsubsec:inference_latency}

Real-time execution is critical for closed-loop autonomous driving, where perception, planning, and control must operate within bounded computation cycles~\citep{weng2024paradrive,fang2020dynamicdeadlines,lead2025}.
In practical autonomous driving systems, observation, inference, and actuation take time.
The command applied at time $t$ cannot be computed from the observation captured at the same instant, because sensor transmission, neural network inference, planning, control conversion, and actuator communication all introduce non-zero latency.
As a result, the vehicle may execute a command derived from an earlier observation.
In closed-loop driving, this delay is not merely a computational detail: it changes the state reached by the vehicle, affects future observations, and may amplify errors over long horizons.

\begin{figure}[!t]
\vspace{-8pt}
\centering
\includegraphics[width=1\linewidth]{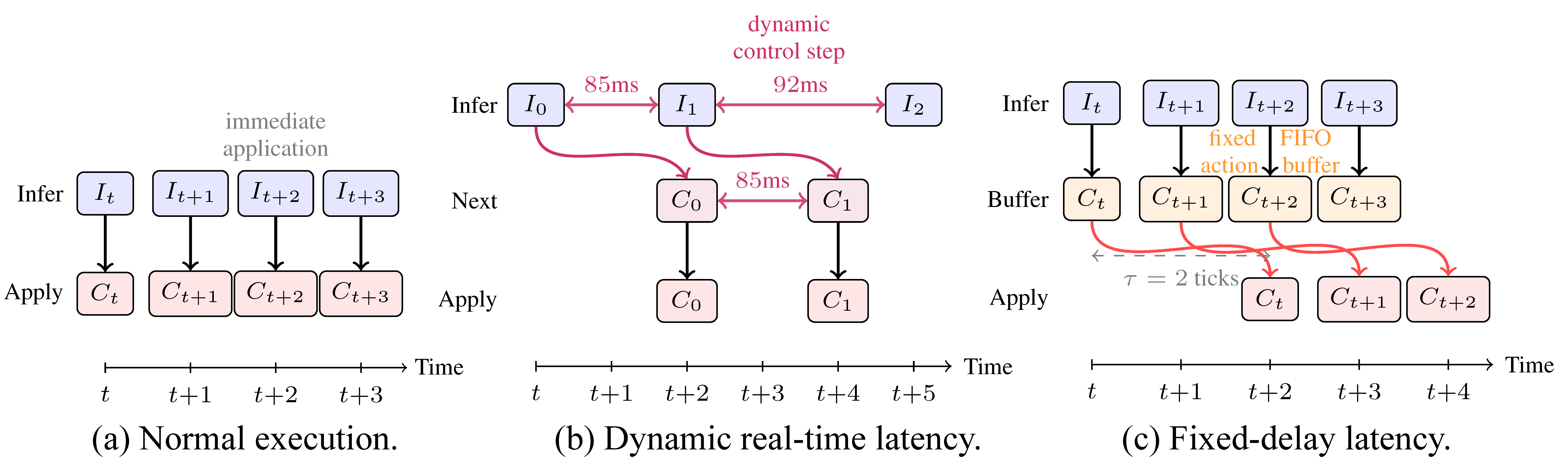}
\vspace{-5pt}
\caption{
\textbf{Latency modes supported in our framework:} immediate execution, dynamic real-time scheduling, and fixed-delay FIFO buffering.
}
\label{fig:latency_comparison}
\vspace{-10pt}
\end{figure}

We therefore model compute-control delay as a FIFO action buffer, as illustrated in Figure~\ref{fig:latency_comparison}.
At each simulation tick, the model computes a control command from the latest available observation, but the command is inserted into the buffer instead of being applied immediately.
The simulator then applies the command generated $\tau$ ticks earlier:
\begin{equation}
\tilde{a}_t = a_{t-\tau}.
\end{equation}
This buffer-based design gives precise control over the latency level, keeps the policy code unchanged, and isolates the robustness effect of delayed actuation from unrelated runtime variations such as GPU load or simulator speed.
Our framework also supports a dynamic real-time latency mode that schedules actions according to measured wall-clock inference time, but fixed-delay latency is used for the main benchmark to ensure comparable severity across models and machines.

\subsubsection{Ego-State Noise}
\label{subsubsec:ego_state_noise}

Ego-state signals provide auxiliary vehicle-state information to the driving policy, such as GPS readings and ego speed.
In real driving environments, GPS localization can be affected by multipath effects, signal blockage, and degraded satellite visibility, and prior studies report meter-level to tens-of-meters localization errors in urban environments~\citep{cui2003gpsurban,brosh2019accurate,rehrl2021localization,gupta2024reliable}.
Vehicle-state sensing can also be affected by sensor and ego-motion reliability issues~\citep{matos2024sensorfailure}.
Motivated by these observations, we evaluate whether E2E-AD policies remain stable when ego-state signals are noisy rather than perfectly clean.

For GPS input noise, we add Gaussian perturbations to the clean GPS reading at each timestep:
\begin{equation}
\tilde{g}_t = g_t + \epsilon_t,
\qquad
\epsilon_t \sim \mathcal{N}(0,\sigma_{\mathrm{GPS}}^2 I),
\end{equation}
where $\sigma_{\mathrm{GPS}}\in\{5,15\}$ m controls the noise level.
This models independent per-timestep GPS input uncertainty rather than a temporally accumulated localization bias.

\begin{figure}[!t]
    \centering
    \includegraphics[width=1\linewidth]{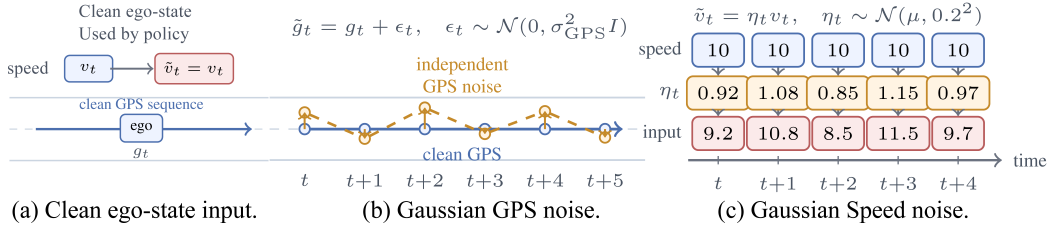}
    \caption{
    \textbf{Illustration of ego-state input perturbations.}
    (a) The model receives the clean GPS reading $g_t$.
    (b) GPS input noise adds Gaussian perturbations to GPS readings.
    (c) Speed noise independently samples a multiplicative factor $\eta_t$ at each timestep and feeds $\tilde{v}_t=\eta_t v_t$ to the policy.
    }
    \label{fig:ego_state_noise}
\end{figure}

For speed noise, we perturb the clean ego speed using an independently sampled multiplicative factor:
\begin{equation}
\tilde{v}_t = \eta_t v_t,
\qquad
\eta_t \sim \mathcal{N}(\mu,0.2^2),
\end{equation}
where $\mu<1$ corresponds to systematic speed underestimation. 
In this benchmark, we focus on underestimation settings with $\mu\in\{0.2,0.5\}$, which model degraded or biased speed sensing.
Figure~\ref{fig:ego_state_noise} summarizes these ego-state perturbations.

        \vspace{-5pt}

\subsubsection{Camera-Stream Perturbations}
\label{subsubsec:camera_stream}

Camera-stream perturbations model failures in how visual observations are captured, transmitted, or delivered to the policy.
Camera crash, frame loss, and sensor corruption have been studied in recent robustness benchmarks and sensor-failure analyses~\citep{xie2025robobev,robodrivebench2025,kong2024robodrive,matos2024sensorfailure}.
We consider two representative camera-stream perturbations: burst frame drop and partial observation.

\begin{figure*}[t]
    \centering
    \includegraphics[width=1\linewidth]{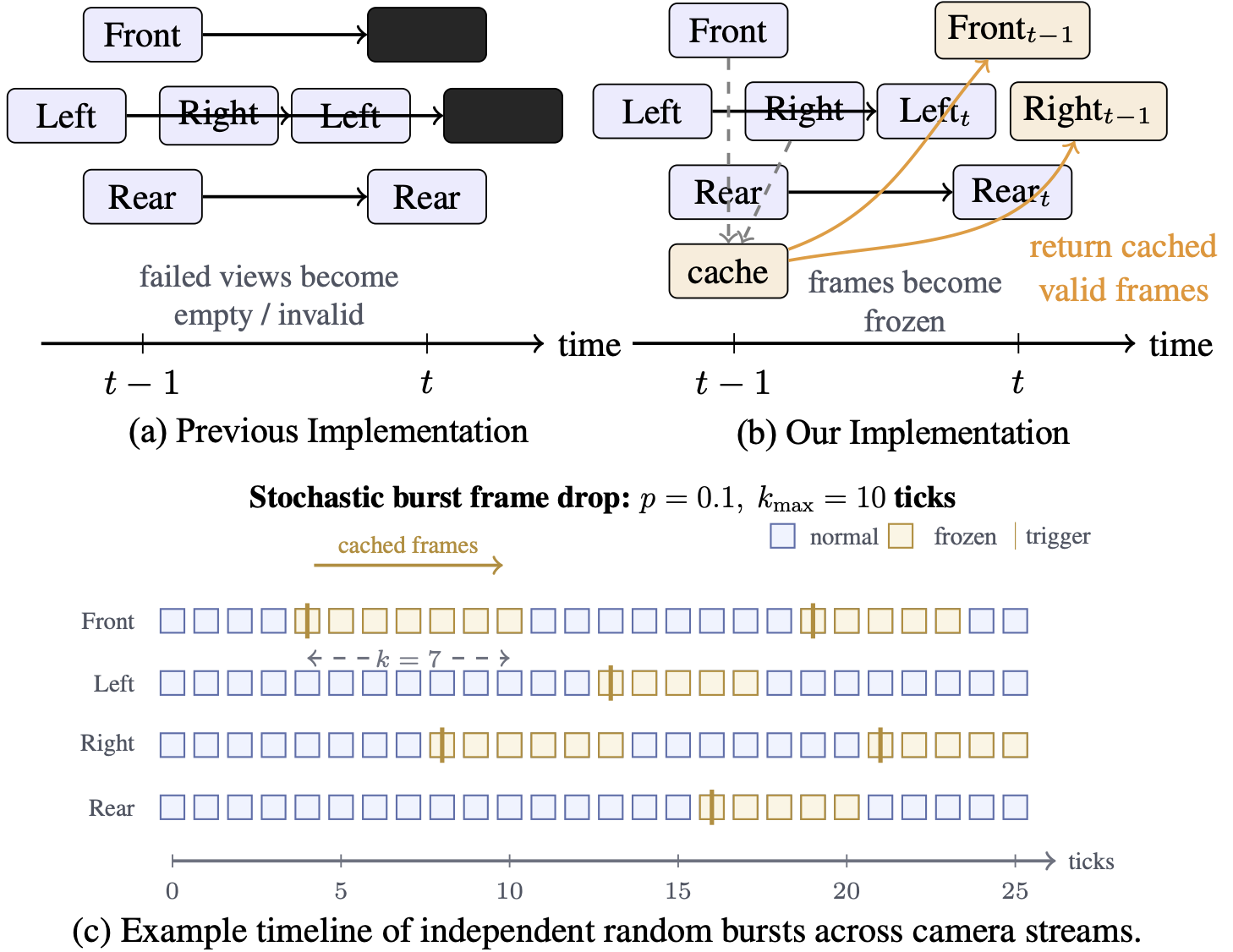}

    \caption{
        \textbf{Illustration of our burst frame drop implementation.}
        (a) Camera crash or frame-lost perturbations typically replace failed views with empty or invalid images.
        (b) In our implementation, a failed camera stream returns the most recent valid cached frame, so the simulator timestamp advances while the visual content is temporally frozen.
        (c) The timeline shows an illustrative example of independently sampled burst events across camera streams with trigger probability $p=0.1$; in the main experiments, burst severity is controlled by the evaluated burst length settings reported in Table~\ref{tab:perturbation_summary}.
        }
    \label{fig:burst_combined}
    \vspace{-10pt}
\end{figure*}

\noindent\textbf{Burst frame drop.}
Unlike camera crash or frame-lost settings that replace the current view with an empty, blacked-out image, our burst frame drop is designed to model a frozen but visually plausible camera stream.
During a burst interval, the affected camera repeatedly returns the most recent valid frame while the simulation timestamp continues to advance.
The delivered image is therefore
\begin{equation}
\tilde{I}_t =
\begin{cases}
I_t, & \text{if the camera stream is normal at time } t, \\
I_{t^\star}, & \text{if the camera is in an active burst interval at time } t,
\end{cases}
\end{equation}
where $t^\star<t$ denotes the timestamp of the most recent valid frame before the burst began.
This design captures a distinct deployment failure mode: the visual input remains plausible, but becomes temporally stale.
Figure~\ref{fig:burst_combined} illustrates the difference between camera crash and our burst frame drop.

\noindent\textbf{Partial observation.}
Partial observation simulates incomplete visual input caused by occlusion or camera-view blockage.
For each image, we apply a black rectangular mask to a randomly sampled region.
The severity is controlled by the mask ratio $r$.
To ensure fair comparison, the mask randomness is generated from deterministic route-level seeds, so different models encounter the same occlusion pattern on the same route.

\begin{figure}[!t]
    \centering
    \includegraphics[width=1\linewidth]{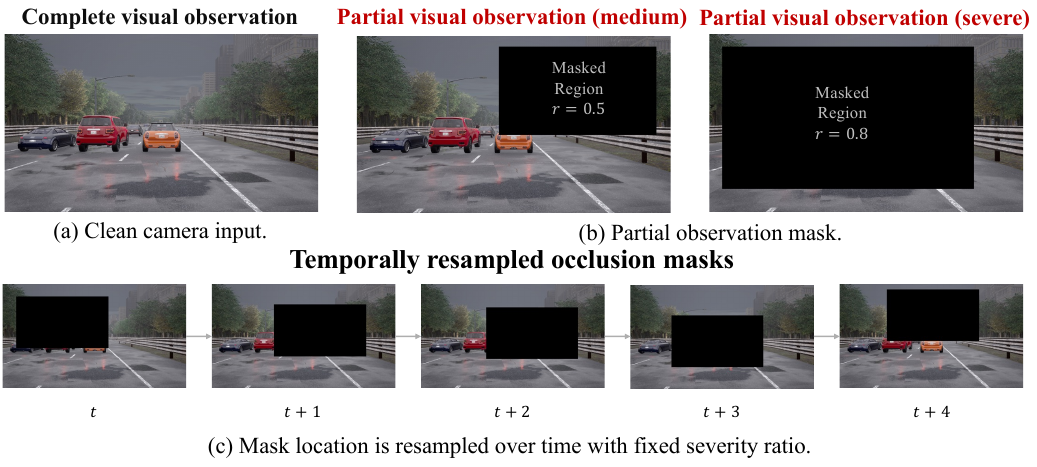}
    \caption{
    \textbf{Illustration of partial observation perturbation.}
    A gray rectangular mask removes part of the camera observation while leaving the external scene unchanged.
    The mask location is resampled over time, and the severity is controlled by the mask ratio $r$.
    }
    \label{fig:partial_observation}
\end{figure}

\subsubsection{Severity Parameterization}
\label{subsec:severity}

Following prior robustness benchmarks, we evaluate perturbations at controlled severity levels to compare models under matched stress conditions~\citep{hendrycks2018benchmarking,dong2023benchmarking,xie2025robobev,jia2024bench2drive}.
Each perturbation type is evaluated at representative severity levels chosen to cover moderate and severe deployment imperfections while keeping the evaluation comparable across models.
The same severity protocol is applied to all evaluated methods.
The complete settings are summarized in Table~\ref{tab:perturbation_summary}.
\subsection{Closed-loop Evaluation Protocol}
We evaluate each driving model in closed-loop simulation under both clean and 
perturbed conditions. For each route and scenario, we first measure clean 
driving performance and then re-evaluate the same model under different 
perturbation categories and severity levels. All evaluated models use the same 
220 Bench2Drive routes, simulator configuration, and perturbation protocol.

This setting allows us to attribute performance changes directly to robustness 
differences rather than to route selection or environment randomness. Because 
the evaluation is performed in closed loop, even small perturbations may 
accumulate over time and manifest as route deviation, unstable control, traffic 
violations, or task failure.

\subsection{Metrics}
\label{subsec:metrics}

We report both standard driving metrics and robustness-oriented degradation 
statistics.

 \noindent\textbf{Standard closed-loop metrics.}
These include Driving Score (DS),  Success Rate (SR), Efficiency (Eff.), and Comfortness (Comf.) inherited from Bench2Drive~\citep{jia2024bench2drive}.

 \noindent\textbf{Robustness degradation.}
We quantify robustness degradation (RD) as the relative performance drop 
compared to the clean baseline:
\begin{equation}
\text{RD}(\pi, \mathcal{P}, s) = 1 - \frac{\text{DS}(\pi, \mathcal{P}, s)}{\text{DS}_\text{clean}(\pi)},
\end{equation}
where $\mathcal{P}$ denotes the perturbation type, $s$ the severity level, 
and $\text{DS}_\text{clean}(\pi)$ the driving score of policy $\pi$ under 
clean conditions. $\text{RD} = 0$ indicates no degradation; $\text{RD} = 1$ 
indicates complete failure.

\section{Experiments}
\label{sec:experiments}

\subsection{Experimental Setup}
\label{subsec:exp_setup}

 \noindent\textbf{Dataset.}
We evaluate Bench2Drive-Robust using the official Bench2Drive closed-loop evaluation protocol, which requires E2E-AD models to complete 220 CARLA routes spanning 44 interactive scenarios under different locations and weather conditions~\citep{jia2024bench2drive}.
Each model is evaluated in closed loop, where its actions affect future states and observations.

 \noindent\textbf{Baselines.}
We evaluate four E2E-AD models: TCP-traj~\citep{wu2022tcp}, UniAD~\citep{hu2023uniad}, VAD~\citep{jiang2023vad}, and SimLingo~\citep{renz2025simlingo}.
All are tested as fixed policies, without finetuning or robustness adaptation.

\noindent\textbf{Perturbations.}
We evaluate camera-stream, ego-state, and compute-control perturbations.
Camera perturbations include partial observation with $r\in\{0.5,0.8\}$ and cached burst frame drop with burst lengths of 20 and 60 ticks, corresponding to 1 and 3 seconds at 20 Hz.
Ego-state perturbations include GPS localization noise with $\sigma_{\mathrm{GPS}}\in\{5\mathrm{m},15\mathrm{m}\}$ and Gaussian multiplicative speed noise $\tilde{v}_t = \eta_t v_t$, where $\eta_t \sim \mathcal{N}(\mu,0.2^2)$.
For SimLingo, GPS perturbations affect the released closed-loop wrapper indirectly by perturbing the GNSS signal used to construct target-point navigation tokens, rather than by feeding raw GPS values directly to the core policy.
Compute-control perturbation is evaluated using fixed inference latency of 100ms, 200ms, and 500ms with a FIFO action buffer.

\subsection{Results}
\label{subsec:exp_results}

\begin{table*}[!t]
\centering
\caption{
\textbf{Closed-loop robustness evaluation on Bench2Drive-Robust.}
We evaluate four E2E-AD models under camera-stream, ego-state, and compute-control perturbations using official Bench2Drive metrics.
RD denotes relative Driving Score degradation from each model's clean baseline.
\textbf{For cross-model robustness comparison, we emphasize RD over absolute DS.}
Speed-N($\mu$) denotes multiplicative speed noise with $\eta\sim\mathcal{N}(\mu,0.2^2)$.
Gray rows are clean baselines.
Avg. perturb., Avg. latency, and Avg. all perturb. average completed non-latency, latency, and all perturbation settings.
}
\label{tab:robust_full_combined}
\scriptsize
\setlength{\tabcolsep}{2.0pt}
\renewcommand{\arraystretch}{1.02}

\begin{minipage}[t]{0.49\textwidth}
\centering
\begin{tabular}{l|l|ccccc}
\hline
\textbf{Method} & \textbf{Setting}
& \textbf{RD} & DS & SR & Eff. & Comf. \\
\hline

\multirow{15}{*}{TCP-traj~\citep{wu2022tcp}}
& \textcolor{gray!50}{baseline}
& \textcolor{gray!50}{0.00}
& \textcolor{gray!50}{59.90}
& \textcolor{gray!50}{30.00}
& \textcolor{gray!50}{76.54}
& \textcolor{gray!50}{18.08} \\
\cline{2-7}
& Occlusion 0.5 & 0.16 & 50.16 & 24.09 & 75.28 & 30.24 \\
& Occlusion 0.8 & 0.24 & 45.54 & 18.18 & 76.10 & 25.24 \\
& Burst 1s & 0.01 & 59.17 & 28.64 & 78.41 & 23.84 \\
& Burst 3s & 0.05 & 56.80 & 25.00 & 78.26 & 23.67 \\
& GPS 5 m & 0.30 & 41.88 & 18.18 & 71.37 & 20.04 \\
& GPS 15 m & 0.65 & 21.12 & 0.45 & 58.37 & 29.22 \\
& Speed-N(0.5) & 0.04 & 57.49 & 27.72 & 91.43 & 23.82 \\
& Speed-N(0.2) & 0.07 & 55.61 & 23.64 & 123.01 & 23.73 \\
& \textbf{Avg. perturb.} & \textbf{0.19} & \textbf{48.47} & \textbf{20.74} & \textbf{81.53} & \textbf{24.98} \\[1.5pt]
\cline{2-7}
& Latency 100 ms & -0.02 & 60.91 & 32.27 & 77.75 & 13.49 \\
& Latency 200 ms & 0.44 & 33.43 & 0.00 & 77.22 & 2.35 \\
& Latency 500 ms & 0.46 & 32.22 & 0.00 & 80.22 & 3.03 \\
& \textbf{Avg. latency} & \textbf{0.30} & \textbf{42.19} & \textbf{10.76} & \textbf{78.40} & \textbf{6.29} \\
\cline{2-7}
& \textbf{Avg. all perturb.} & \textbf{0.22} & \textbf{46.76} & \textbf{18.02} & \textbf{80.67} & \textbf{19.88} \\[2pt]
\hline

\multirow{15}{*}{UniAD~\citep{hu2023uniad}}
& \textcolor{gray!50}{baseline}
& \textcolor{gray!50}{0.00}
& \textcolor{gray!50}{45.81}
& \textcolor{gray!50}{16.36}
& \textcolor{gray!50}{129.21}
& \textcolor{gray!50}{43.58} \\
\cline{2-7}
& Occlusion 0.5 & 0.11 & 40.71 & 15.45 & 127.46 & 42.25 \\
& Occlusion 0.8 & 0.39 & 28.03 & 2.27 & 133.95 & 46.66 \\
& Burst 1s & 0.03 & 44.26 & 14.55 & 134.70 & 27.50 \\
& Burst 3s & 0.08 & 42.20 & 13.64 & 126.88 & 43.82 \\
& GPS 5 m & 0.15 & 39.07 & 11.36 & 141.52 & 51.11 \\
& GPS 15 m & 0.55 & 20.50 & 0.00 & 117.55 & 48.91 \\
& Speed-N(0.5) & 0.19 & 37.27 & 12.72 & 184.92 & 51.85 \\
& Speed-N(0.2) & 0.37 & 29.08 & 4.55 & 221.40 & 54.46 \\
& \textbf{Avg. perturb.} & \textbf{0.23} & \textbf{35.14} & \textbf{9.32} & \textbf{148.55} & \textbf{45.82} \\[1.5pt]
\cline{2-7}
& Latency 100 ms & 0.19 & 37.01 & 14.09 & 123.44 & 47.00 \\
& Latency 200 ms & 0.26 & 33.84 & 10.45 & 130.93 & 8.12 \\
& Latency 500 ms & 0.30 & 31.85 & 7.73 & 143.88 & 5.42 \\
& \textbf{Avg. latency} & \textbf{0.25} & \textbf{34.23} & \textbf{10.76} & \textbf{132.75} & \textbf{20.18} \\
\cline{2-7}
& \textbf{Avg. all perturb.} & \textbf{0.24} & \textbf{34.89} & \textbf{9.71} & \textbf{144.24} & \textbf{38.83} \\
\hline
\end{tabular}
\end{minipage}
\hfill
\begin{minipage}[t]{0.49\textwidth}
\centering
\begin{tabular}{l|l|ccccc}
\hline
\textbf{Method} & \textbf{Setting}
& \textbf{RD} & DS & SR & Eff. & Comf. \\
\hline

\multirow{15}{*}{VAD~\citep{jiang2023vad}}
& \textcolor{gray!50}{baseline}
& \textcolor{gray!50}{0.00}
& \textcolor{gray!50}{42.35}
& \textcolor{gray!50}{15.00}
& \textcolor{gray!50}{157.94}
& \textcolor{gray!50}{46.01} \\
\cline{2-7}
& Occlusion 0.5 & 0.00 & 42.21 & 17.73 & 163.26 & 41.88 \\
& Occlusion 0.8 & 0.22 & 33.15 & 6.39 & 167.09 & 49.50 \\
& Burst 1s & -0.02 & 43.04 & 17.73 & 151.75 & 47.69 \\
& Burst 3s & 0.08 & 39.09 & 13.18 & 149.70 & 50.76 \\
& GPS 5 m & 0.13 & 37.02 & 12.27 & 163.82 & 32.64 \\
& GPS 15 m & 0.62 & 16.26 & 0.00 & 155.12 & 57.99 \\
& Speed-N(0.5) & 0.06 & 39.64 & 14.09 & 213.24 & 50.31 \\
& Speed-N(0.2) & 0.19 & 34.33 & 8.64 & 243.92 & 54.21 \\
& \textbf{Avg. perturb.} & \textbf{0.16} & \textbf{35.59} & \textbf{11.25} & \textbf{175.99} & \textbf{48.12} \\[1.5pt]
\cline{2-7}
& Latency 100 ms & 0.14 & 36.52 & 11.36 & 144.35 & 53.10 \\
& Latency 200 ms & 0.36 & 27.15 & 3.64 & 252.31 & 11.70 \\
& Latency 500 ms & 0.44 & 23.55 & 2.27 & 242.62 & 9.75 \\
& \textbf{Avg. latency} & \textbf{0.31} & \textbf{29.07} & \textbf{5.76} & \textbf{213.09} & \textbf{24.85} \\
\cline{2-7}
& \textbf{Avg. all perturb.} & \textbf{0.20} & \textbf{33.81} & \textbf{9.75} & \textbf{186.11} & \textbf{41.78} \\[2pt]
\hline

\multirow{15}{*}{SimLingo~\citep{renz2025simlingo}}
& \textcolor{gray!50}{baseline}
& \textcolor{gray!50}{0.00}
& \textcolor{gray!50}{85.94}
& \textcolor{gray!50}{66.82}
& \textcolor{gray!50}{244.18}
& \textcolor{gray!50}{25.49} \\
\cline{2-7}
& Occlusion 0.5 & 0.29 & 60.71 & 25.00 & 226.06 & 37.57 \\
& Occlusion 0.8 & 0.83 & 14.79 & 0.00 & 199.93 & 70.22 \\
& Burst 1s & 0.01 & 85.47 & 66.36 & 235.40 & 31.01 \\
& Burst 3s & 0.00 & 85.82 & 69.09 & 236.16 & 31.68 \\
& GPS 5 m & -0.02 & 87.91 & 73.64 & 238.02 & 33.41 \\
& GPS 15 m & -0.02 & 87.53 & 70.91 & 238.64 & 32.98 \\
& Speed-N(0.5) & 0.28 & 61.77 & 29.09 & 276.62 & 33.51 \\
& Speed-N(0.2) & 0.50 & 42.73 & 6.36 & 282.45 & 27.15 \\
& \textbf{Avg. perturb.} & \textbf{0.23} & \textbf{65.84} & \textbf{42.56} & \textbf{241.66} & \textbf{37.19} \\
\cline{2-7}
& Latency 100 ms & 0.67 & 28.45 & 2.27 & 218.92 & 63.66 \\
& Latency 200 ms & 0.77 & 19.47 & 0.00 & 189.95 & 64.52 \\
& Latency 500 ms & 0.82 & 15.70 & 0.00 & 177.93 & 59.78 \\
& \textbf{Avg. latency} & \textbf{0.75} & \textbf{21.21} & \textbf{0.76} & \textbf{195.60} & \textbf{62.65} \\
\cline{2-7}
& \textbf{Avg. all perturb.} & \textbf{0.38} & \textbf{53.67} & \textbf{31.16} & \textbf{229.10} & \textbf{44.14} \\

\end{tabular}
\end{minipage}

\vspace{-10pt}
\end{table*}

\noindent\textbf{Key findings.}
Table~\ref{tab:robust_full_combined} shows that deployment-side perturbations induce heterogeneous closed-loop failures across perturbation families.
Since Bench2Drive evaluates E2E-AD agents in closed-loop over 220 CARLA routes spanning 44 interactive scenarios, these degradations reflect changes in full driving rollouts rather than isolated frame-level errors~\citep{jia2024bench2drive}.
The qualitative case studies in Appendix~\ref{app:case_study} further illustrate how the aggregate drops translate into concrete route-level failures.

\textbf{Heterogeneous robustness.}
Robustness cannot be inferred from a single clean-condition score.
For example, SimLingo achieves the strongest clean Driving Score, but degrades sharply under severe occlusion and inference latency.
In contrast, several lower-clean-score models are less affected by cached burst frame drop.
This suggests that each perturbation family probes a different failure axis of the closed-loop driving stack, rather than a single generic robustness property.

\textbf{Camera-stream failures.}
Severe occlusion substantially reduces Driving Score and Success Rate by removing task-relevant visual evidence.
The actor-flow case in Fig.~\ref{fig:case_study_occlusion} shows that occlusion can hide nearby actors, free space, and the intended turning corridor, making the interaction decision unreliable.
Cached burst frame drop is generally less harmful for TCP-traj, UniAD, and VAD, suggesting that stale but visually plausible frames can be less destructive than direct spatial information loss.
However, burst drops still break temporal consistency between the delivered image and the evolving scene.

\textbf{Ego-state sensitivity.}
GPS and speed perturbations show that robustness failures are not limited to camera inputs.
The GPS case in Fig.~\ref{fig:case_study_gps_tcp} shows how localization noise can misalign the route target and predicted waypoints, causing the trajectory-following controller to deviate from the intended turn.
The speed case in Fig.~\ref{fig:case_study_speed} shows a complementary longitudinal-control failure: severe speed underestimation leads to insufficient braking at a red light and unstable behavior during the subsequent turn.
These results indicate that auxiliary state inputs can become critical failure points even when visual observations remain available.

\begin{figure}[!t]
    \centering
    \includegraphics[width=1\linewidth]{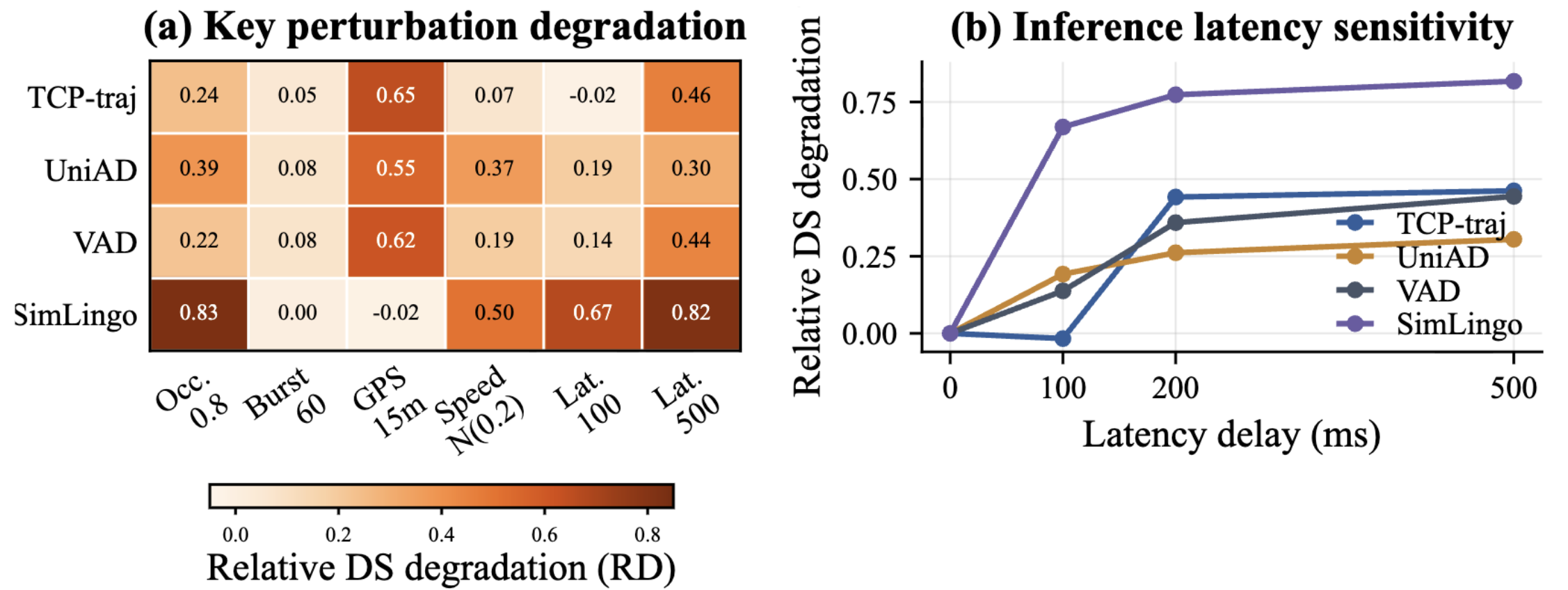}
    \caption{
    \textbf{Main robustness analysis.}
    Deployment-side perturbations induce heterogeneous degradation patterns across models, and inference latency reveals strong closed-loop synchronization vulnerability.
    }
    \label{fig:main_analysis}
    \vspace{-10pt}
\end{figure}

\textbf{Latency vulnerability.}
Inference latency exposes a strong synchronization failure mode.
All methods degrade as delay increases, and SimLingo drops sharply despite having the strongest clean baseline.
As shown in Fig.~\ref{fig:case_study_latency}, predicted waypoints may remain visually plausible, but delayed execution makes the corresponding action arrive after the scene has evolved.
Thus, latency primarily breaks perception--action timing rather than necessarily producing obviously invalid predictions.

\textbf{Metric interpretation.}
We treat DS, RD, and SR as the primary robustness indicators because they most directly reflect closed-loop task completion and safety-related degradation.
Efficiency and Comfortness are useful complementary diagnostics, but they can be non-monotonic under failed or shortened rollouts.
The case studies support this interpretation: failures such as collision, blocking, delayed execution, and insufficient braking are most clearly captured by Driving Score degradation and Success Rate changes.

\section{Conclusion}

We present \textbf{Bench2Drive-Robust}, a closed-loop robustness benchmark for E2E-AD under deployment-oriented perturbations.
It injects camera-stream failures, ego-state noise, speed sensing errors, and inference latency into Bench2Drive while keeping the evaluated policy unchanged.
Experiments show that these perturbations substantially degrade closed-loop performance and that clean-driving performance does not necessarily imply robustness.
Our benchmark provides a diagnostic protocol for future deployment-aware E2E-AD systems.


\bibliographystyle{unsrtnat}   
\bibliography{main}

\clearpage

\appendix

\section{Limitations}
\label{app:limitations}
Bench2Drive-Robust is built on CARLA-based closed-loop simulation, which enables reactive and repeatable evaluation under structured perturbations.
However, simulation rendering and traffic behavior may differ from real-world driving distributions.
Real-world dataset-based benchmarks provide complementary visual realism, but are typically non-reactive and therefore cannot fully capture closed-loop feedback.
Future work may combine reactive simulation with real-world-calibrated perturbation statistics or generative simulation to improve both realism and interactivity.

\section{Implementation Details}
\label{app:implementation}

\subsection{Perturbation Injection Architecture}
\label{app:injection_architecture}

\begin{figure*}[t]
    \centering
    \includegraphics[width=0.92\textwidth]{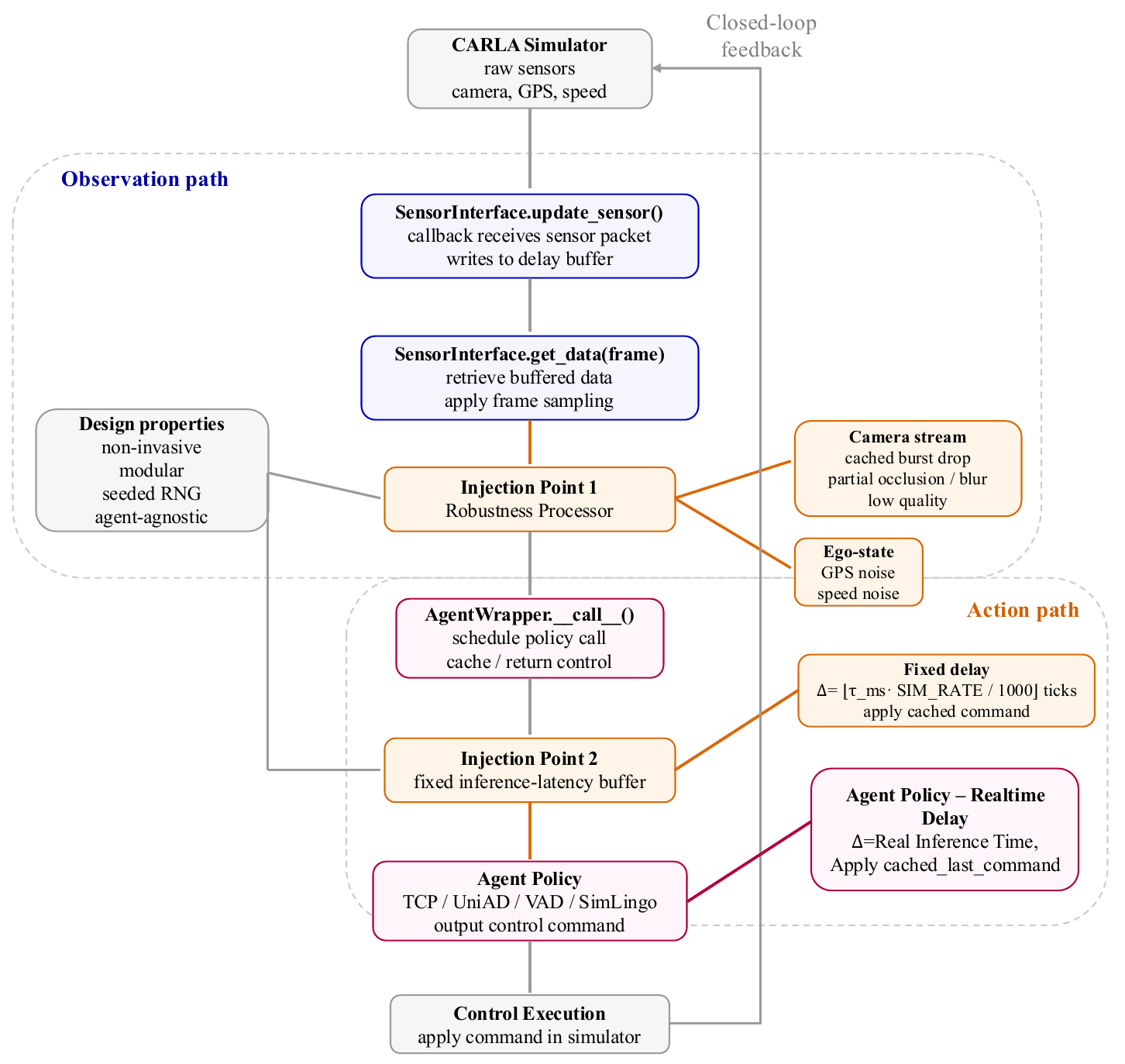}
    \caption{
    \textbf{Perturbation injection architecture.}
    Bench2Drive-Robust injects observation-side perturbations before policy inference and action-side latency before command execution, while keeping the evaluated model unchanged.
    }
    \label{fig:injection_architecture}
    \vspace{-6pt}
\end{figure*}

Figure~\ref{fig:injection_architecture} shows the two non-invasive injection points used by Bench2Drive-Robust.
The first injection point lies on the observation path.
CARLA sensor packets are received by \texttt{SensorInterface.update\_sensor()} through sensor callbacks and stored in a delay buffer.
When the model requests observations through \texttt{SensorInterface.get\_data(frame)}, the interface retrieves buffered measurements, applies the original frame-sampling logic, and then passes the result to the robustness processor.
This processor applies camera-stream perturbations, including cached burst frame drop, partial occlusion, as well as ego-state perturbations such as GPS noise and multiplicative speed noise.
Since these transformations are applied before policy inference but after sensor retrieval, the original model input interface is preserved.

The second injection point lies on the action path.
Each evaluation step invokes \texttt{AgentWrapper.\_\_call\_\_()}, where the wrapper either calls the underlying policy normally or returns a delayed/cached control command.
For fixed latency, the target delay is converted into simulation ticks as
\begin{equation}
\Delta=\left\lfloor \tau_{\mathrm{ms}}\cdot \mathrm{SIM\_RATE}/1000 \right\rfloor .
\end{equation}
For real-time delay, the wrapper uses measured inference time and reuses the last cached command when the policy misses the control cycle.
Both modes emulate delayed action execution without changing the policy architecture, checkpoint, or inference code.

This implementation has three practical advantages.
First, it is model-agnostic: the same perturbation interface can be applied to TCP-traj, UniAD, VAD, SimLingo, or other Bench2Drive-compatible agents.
Second, it is modular and reproducible: each perturbation is implemented as an independent processor and stochastic operations are controlled by seeded random number generators.
Third, it separates observation corruption from action delay, enabling camera-stream failures, ego-state errors, and compute-control latency to be evaluated independently or in combination under the same closed-loop protocol.

\subsection{Reproducibility and Configuration}
\label{app:reproducibility_config}

Table~\ref{tab:app_config} summarizes the main configuration variables used to control Bench2Drive-Robust.
We group the variables by perturbation type to make the implementation protocol explicit: global variables enable robustness evaluation and fix random seeds; camera-stream variables control cached burst frame drop and partial observation; ego-state variables control GPS and speed perturbations; and compute-control variables specify inference-latency injection.
All perturbations are configured externally through environment variables rather than by modifying the evaluated model, which keeps the benchmark model-agnostic and makes repeated evaluations reproducible under the same route, seed, and severity setting.

\begin{table}[t]
\centering
\caption{Main configuration variables used by Bench2Drive-Robust.}
\label{tab:app_config}
\small
\setlength{\tabcolsep}{4pt}
\renewcommand{\arraystretch}{1.08}
\begin{tabularx}{\linewidth}{l l X}
\toprule
\textbf{Perturbation Type} & \textbf{Variable} & \textbf{Description} \\
\midrule

\multirow{2}{*}{Global}
& \texttt{ROBUSTNESS\_ENABLE} & Enable observation-side perturbations \\
& \texttt{ROBUSTNESS\_SEED} & Seed for stochastic perturbations \\
\midrule

\multirow{3}{*}{Camera stream}
& \texttt{BURST\_MAX\_TICKS} & Maximum cached burst duration \\
& \texttt{BURST\_PROBABILITY} & Probability of triggering a burst event \\
& \texttt{PARTIAL\_OBS\_RATIO} & Occlusion mask ratio for partial observation \\
\midrule

\multirow{3}{*}{Ego-state}
& \texttt{GPS\_NOISE\_STD} & Standard deviation of GPS localization noise \\
& \texttt{SPEED\_BIAS\_MEAN} & Mean of the Gaussian speed multiplier \\
& \texttt{SPEED\_BIAS\_STD} & Standard deviation of the Gaussian speed multiplier \\
\midrule

\multirow{3}{*}{Compute-control}
& \texttt{INFERENCE\_LATENCY\_ENABLE} & Enable action-side latency injection \\
& \texttt{INFERENCE\_LATENCY\_MS} & Fixed inference latency in milliseconds \\
& \texttt{SIM\_RATE} & Simulation/control rate used to convert milliseconds to ticks \\
\bottomrule
\end{tabularx}
\end{table}

\subsection{Model Configuration}
\label{app:model_config}

All methods are evaluated using their released or Bench2Drive-compatible closed-loop configurations without robustness finetuning.
We keep each agent's architecture, checkpoint, planner logic, and controller unchanged, and apply perturbations only at the simulator--policy interface.
Because models differ in camera setup, navigation interface, and ego-state usage, the same perturbation may affect different methods through different input channels.
Thus, weak sensitivity to GPS or speed perturbations does not necessarily indicate intrinsic robustness; it may reflect whether the released closed-loop wrapper directly consumes that signal.
We therefore interpret the results as deployed-model robustness profiles rather than architecture-normalized comparisons.

\section{Further Analysis}
\label{app:further_analysis}

\begin{figure}[p]
    \centering

    \begin{subfigure}[t]{1\textwidth}
        \centering
        \includegraphics[width=\linewidth]{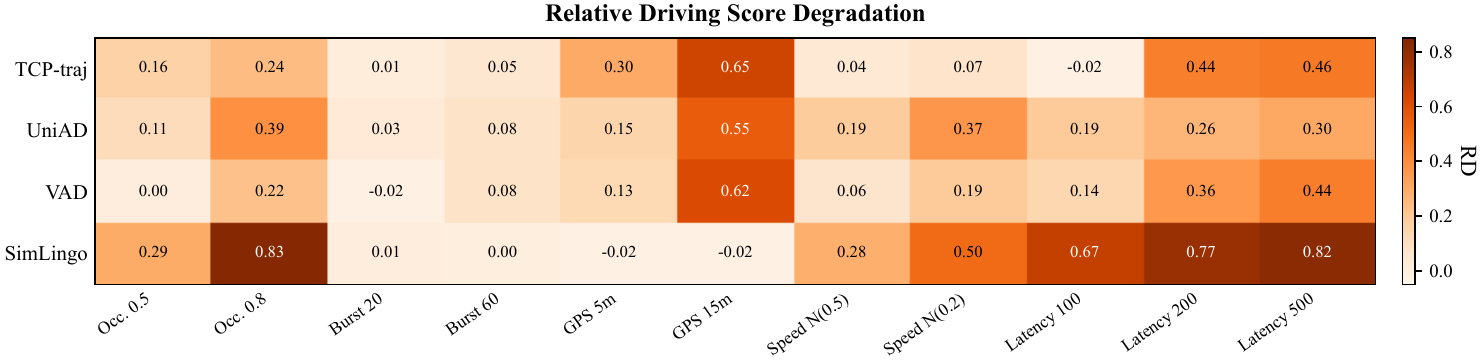}
        \caption{Relative Driving Score degradation across perturbations and models.}
        \label{fig:app_rd_heatmap}
    \end{subfigure}

    \vspace{1.0em}

    \begin{subfigure}[t]{1\textwidth}
        \centering
        \includegraphics[width=\linewidth]{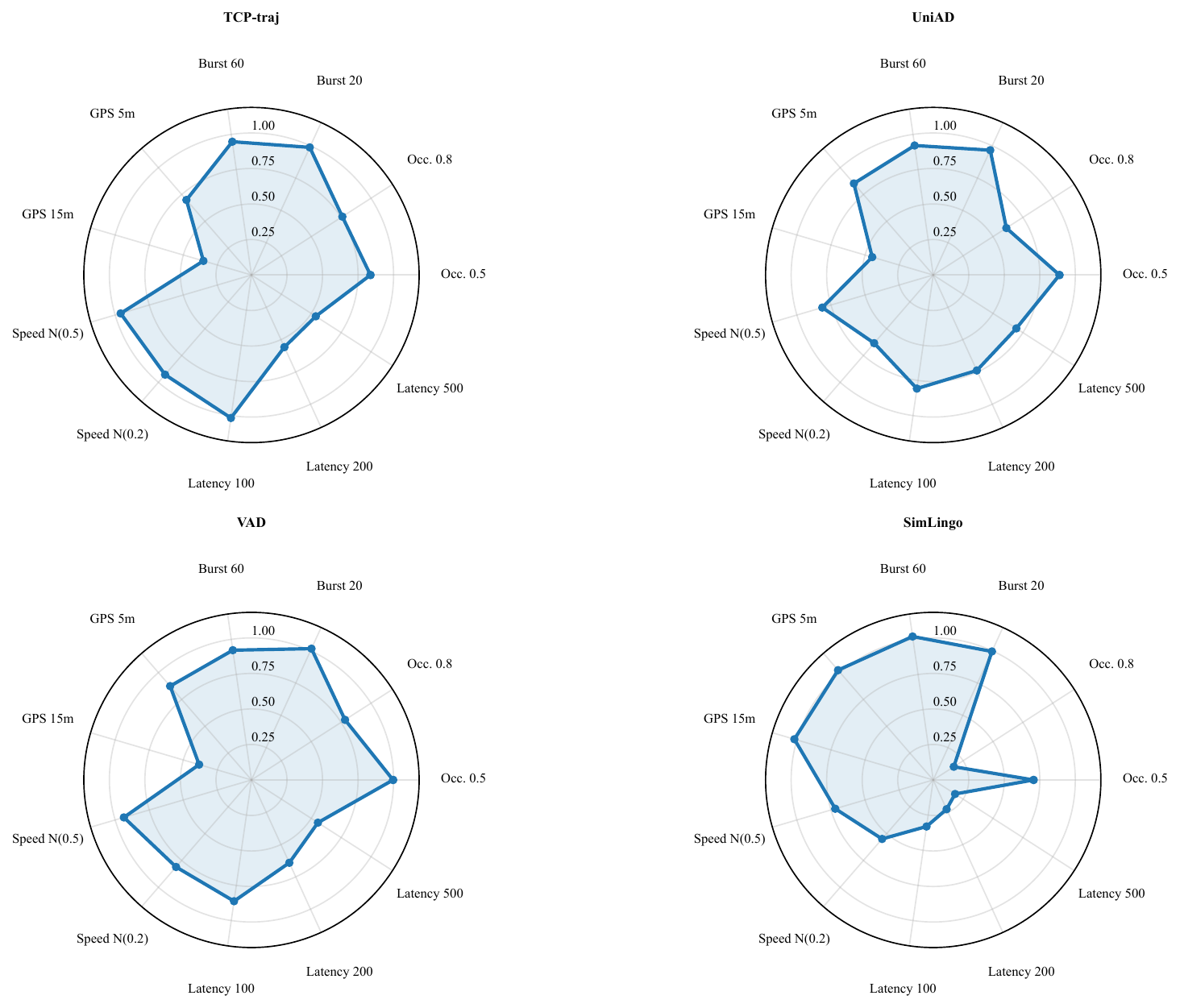}
        \caption{Per-model Driving Score retention profiles across perturbations.}
        \label{fig:app_radar_retention_per_model}
    \end{subfigure}

    \caption{
\textbf{    Overall robustness overview across perturbation types and models.
}    The heatmap reports relative Driving Score degradation with respect to each model's clean baseline, where larger values indicate stronger degradation.
    The radar plots provide a complementary per-model view by showing Driving Score retention, defined as the ratio between perturbed Driving Score and clean baseline Driving Score.
    Together, these visualizations show that robustness behavior is highly heterogeneous across models and perturbation families: methods with strong clean-driving performance can still suffer substantial degradation under specific system-level perturbations such as severe occlusion, GPS localization noise, or inference latency.
    Missing entries are marked as unavailable.
    }
    \label{fig:app_overall_robustness_overview}
\end{figure}

Figure~\ref{fig:app_overall_robustness_overview} provides an overall robustness overview across perturbation types and models.
The heatmap in Fig.~\ref{fig:app_rd_heatmap} summarizes relative Driving Score degradation for each method--perturbation pair, while the radar plots in Fig.~\ref{fig:app_radar_retention_per_model} show the corresponding per-model retention profiles.
Together, these visualizations reveal heterogeneous robustness patterns across models and perturbation families.
TCP-traj maintains relatively high retention under mild camera-side perturbations and speed noise, but its performance drops substantially under severe GPS localization noise and larger inference latency, indicating sensitivity to localization and delayed control feedback.
UniAD and VAD exhibit smoother degradation across most perturbations, but both still suffer clear retention loss under severe GPS localization noise and high-latency settings.
SimLingo achieves the strongest clean baseline performance, yet its robustness profile is highly uneven: the heatmap shows sharp degradation under severe occlusion and inference latency, while the radar profile shows that it retains performance under some available ego-state perturbation settings.
These patterns suggest that clean closed-loop performance alone is insufficient to characterize deployment robustness, since different models fail under different system components and severity regimes.

\begin{figure}[p]
    \centering

    \begin{subfigure}[t]{0.8\textwidth}
        \centering
        \includegraphics[width=\linewidth]{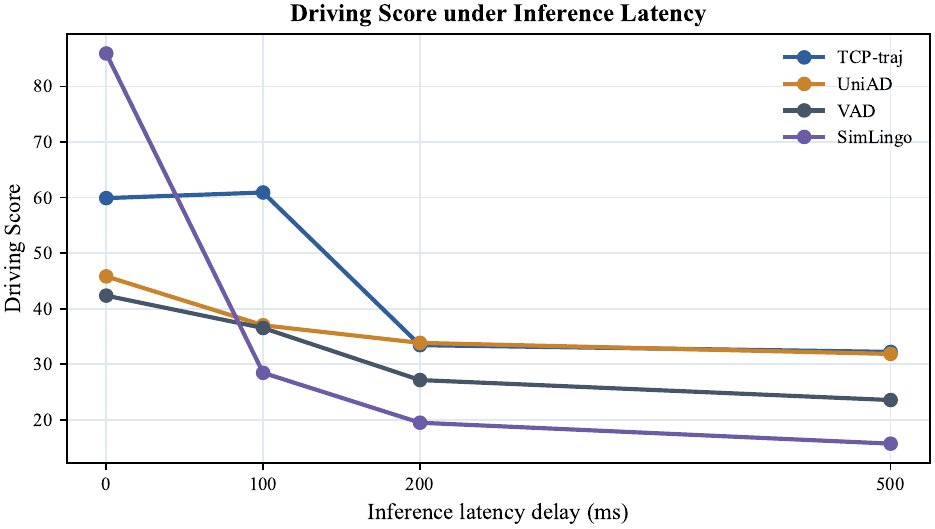}
        \caption{Driving Score under increasing inference latency.}
        \label{fig:app_inference_ds_curve}
    \end{subfigure}

    \vspace{0.8em}

    \begin{subfigure}[t]{0.8\textwidth}
        \centering
        \includegraphics[width=\linewidth]{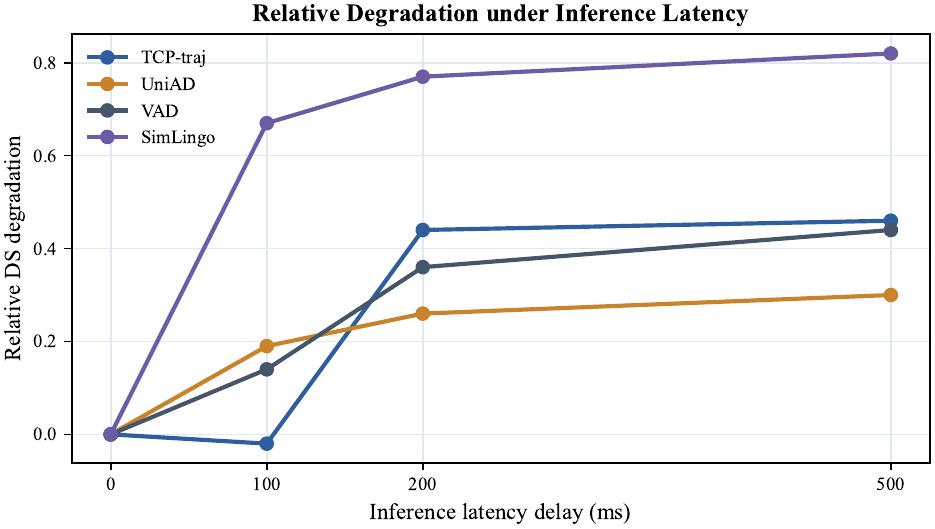}
        \caption{Relative Driving Score degradation under increasing inference latency.}
        \label{fig:app_inference_rd_curve}
    \end{subfigure}

    \vspace{0.8em}

    \begin{subfigure}[t]{0.8\textwidth}
        \centering
        \includegraphics[width=\linewidth]{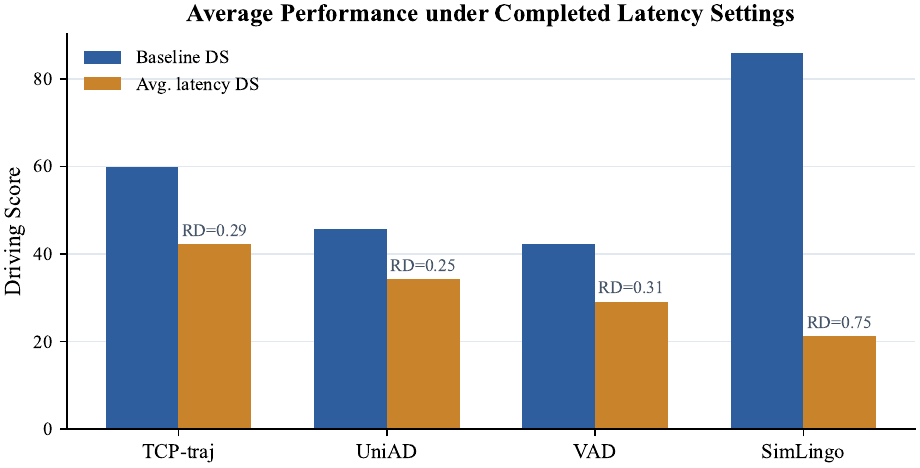}
        \caption{Clean baseline performance versus average performance across latency settings.}
        \label{fig:app_inference_avg_latency}
    \end{subfigure}

    \caption{
    \textbf{Detailed analysis of inference latency robustness under delayed control execution.}
    The first two plots show absolute Driving Score and relative degradation across latency settings of 0ms, 100ms, 200ms, and 500ms.
    The third plot compares each model's clean baseline Driving Score with its average Driving Score under completed latency settings, with annotations showing the average relative degradation.
    Together, these results highlight that latency sensitivity differs substantially across models and that strong clean-driving performance does not necessarily imply robustness to delayed control.
    }
    \label{fig:app_inference_latency_breakdown}
\end{figure}

Figure~\ref{fig:app_inference_latency_breakdown} provides a detailed view of inference latency robustness.
The absolute Driving Score curve shows how each model's closed-loop performance changes as delay increases, while the relative degradation curve normalizes this effect by each model's clean baseline.
The average latency summary further compares nominal performance against average delayed-control performance.
These views show that latency sensitivity differs substantially across models and that strong clean-driving performance does not necessarily imply robustness to delayed control.

\clearpage

\clearpage

\clearpage
\clearpage

\section{Case Study}
\label{app:case_study}
\subsection{Occlusion}
\begin{figure}[t]
    \centering
    \includegraphics[width=1\linewidth]{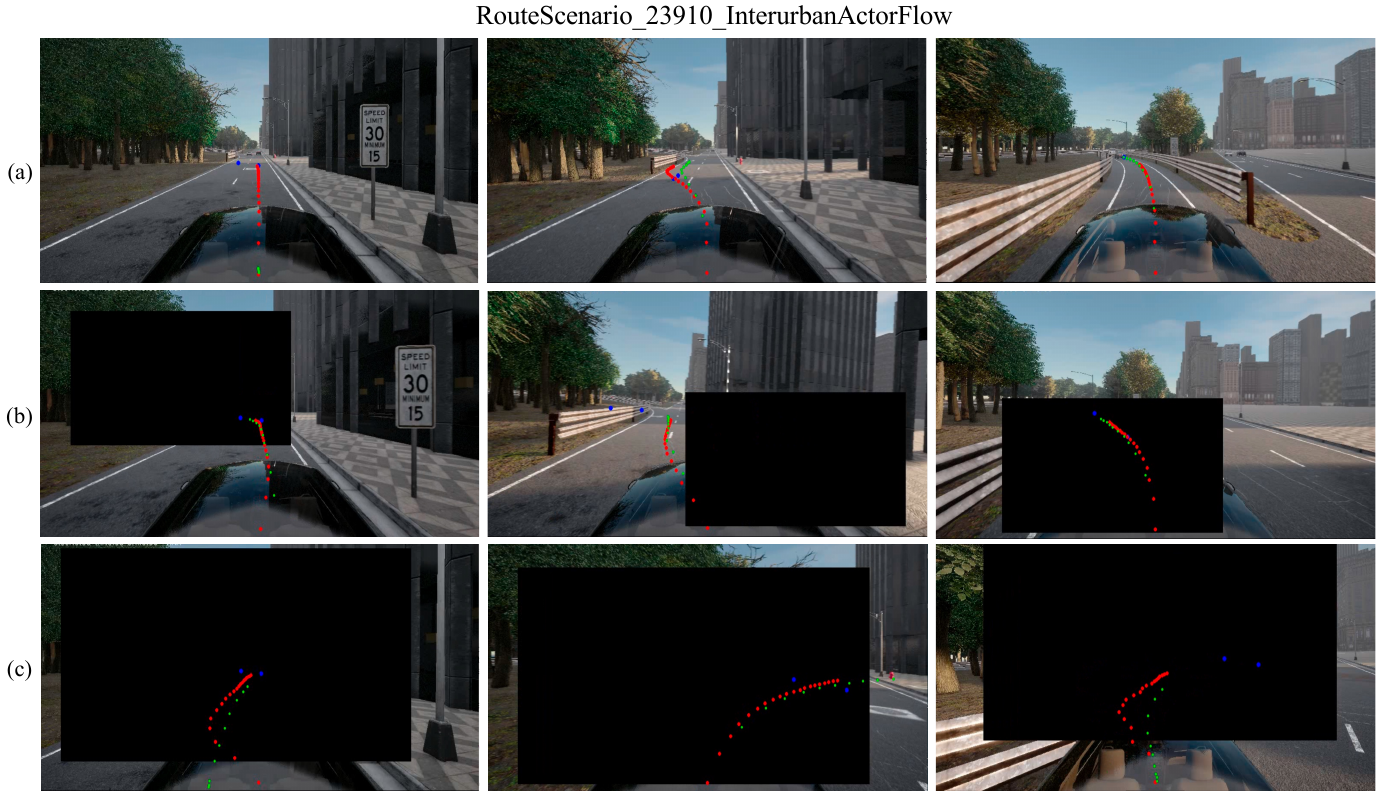}
    \caption{
        \textbf{Qualitative occlusion case study on SimLingo.}
        We compare the same route under three settings:
        (a) clean baseline,
        (b) partial observation with occlusion ratio $r=0.5$,
        and (c) severe partial observation with occlusion ratio $r=0.8$.
        The overlaid colored points follow the SimLingo visualization convention:
        \textcolor{blue}{blue points} denote navigation target points,
        \textcolor{red}{red points} denote predicted path waypoints,
        and \textcolor{green!60!black}{green points} denote predicted speed waypoints.
        The comparison illustrates how progressively stronger visual occlusion removes task-relevant scene evidence and leads to degraded closed-loop behavior.
        }
    \label{fig:case_study_occlusion}
\end{figure}
Figure~\ref{fig:case_study_occlusion} shows an occlusion case study for SimLingo on RouteScenario\_23910\_rep0, which belongs to the \texttt{InterurbanActorFlow} scenario in Bench2Drive.
In this scenario, the ego vehicle leaves an interurban road by making a left turn while crossing a fast traffic flow~\citep{jia2024bench2drive}.
This route requires several coupled driving abilities: recognizing the road geometry and intended left-turn path, perceiving fast-moving surrounding vehicles, estimating safe temporal gaps, and maintaining controlled progress while crossing the traffic stream.
Because the maneuver depends on both global route context and local actor-level evidence, it is particularly sensitive to partial observation.

Under the clean baseline in Fig.~\ref{fig:case_study_occlusion}(a), SimLingo receives complete camera observations and can infer the road layout, nearby vehicles, and available driving corridor.
With moderate occlusion in Fig.~\ref{fig:case_study_occlusion}(b), the mask removes part of the visible scene, making local interaction cues less reliable even though some route context remains visible.
Under severe occlusion in Fig.~\ref{fig:case_study_occlusion}(c), the visible region is substantially reduced, making it harder to infer nearby actors, free space, and the timing of the crossing maneuver.
This example complements the aggregate results in Table~\ref{tab:robust_full_combined}: although SimLingo achieves strong clean performance, its closed-loop behavior can degrade sharply when task-relevant visual evidence is removed.
\clearpage

\clearpage
\subsection{Inference Latency}
\begin{figure}[t]
    \centering
    \includegraphics[width=1\linewidth]{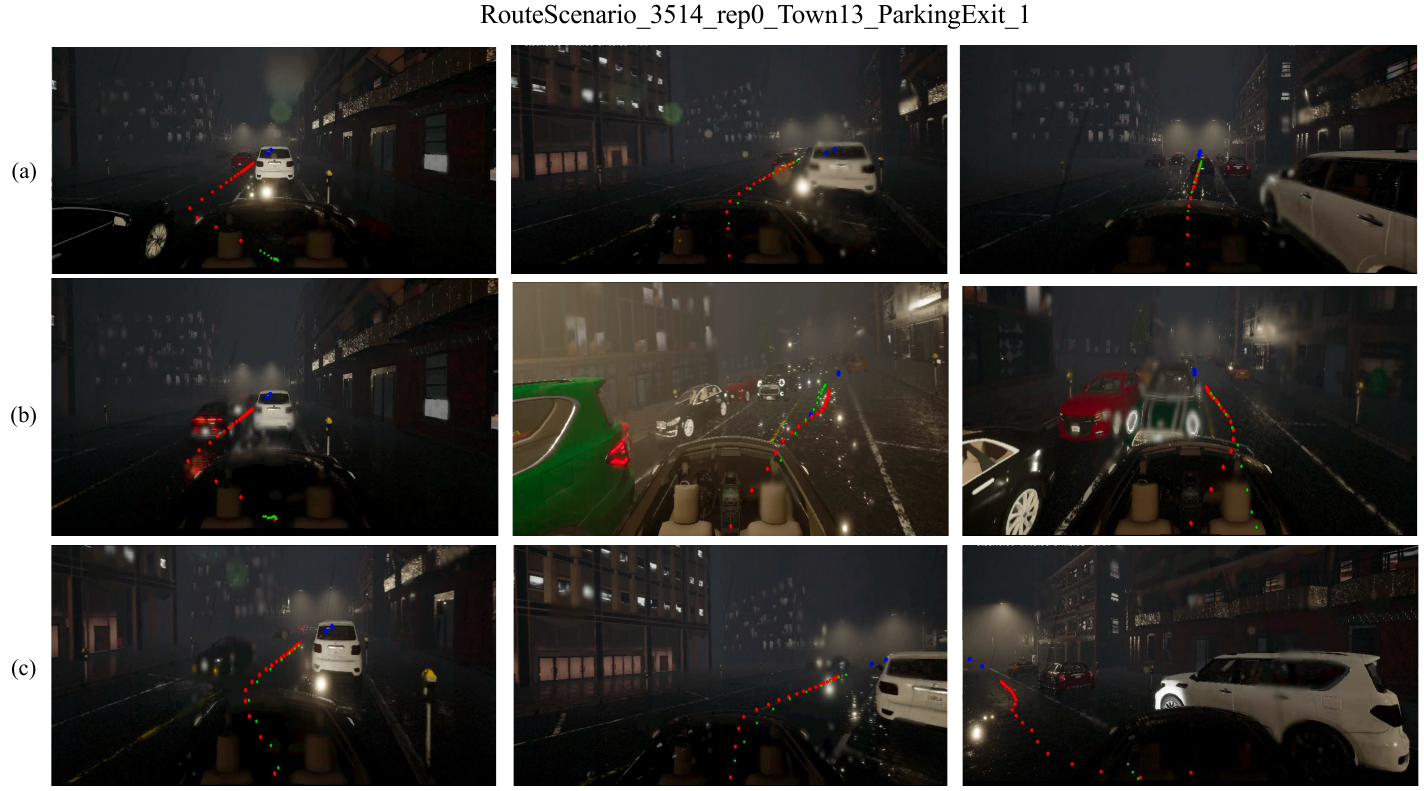}
    \caption{
    \textbf{Qualitative inference-latency case study on SimLingo.}
    We compare the same \texttt{ParkingExit} route under:
    (a) clean execution,
    (b) 100ms inference latency,
    and (c) 500ms inference latency.
    The route requires the ego vehicle to exit a parallel parking bay and merge into traffic with timely steering and acceleration.
    \textcolor{blue}{Blue points} denote navigation target points,
    \textcolor{red}{red points} denote predicted path waypoints,
    and \textcolor{green!60!black}{green points} denote predicted speed waypoints.
    Although the predicted waypoints remain plausible, delayed action execution causes temporally misaligned control and degraded closed-loop behavior.
    }
    \label{fig:case_study_latency}
\end{figure}
Figure~\ref{fig:case_study_latency} shows a qualitative inference-latency case study for SimLingo on a \texttt{ParkingExit} route.
In this scenario, the ego vehicle must exit a parallel parking bay into an active traffic flow, which requires understanding the local road geometry, selecting a feasible exit path, estimating nearby vehicle motion, and executing steering and acceleration \textit{at the correct time}.
This makes the route sensitive to action delay: even if the predicted trajectory is reasonable for the observation time, the command may become outdated by the time it is executed.
The \texttt{ParkingExit} scenario is designed to test this type of interactive maneuver, where the ego vehicle must leave a parallel parking bay and enter a traffic flow~\citep{jia2024bench2drive}.

Figure~\ref{fig:case_study_latency}(a) shows that under clean baseline execution, SimLingo predicts a reasonable waypoint sequence and executes the left-then-right parking-exit maneuver at the correct timing.
In Figure~\ref{fig:case_study_latency}(b), with 100ms latency, the predicted waypoints remain visually plausible, but the delayed execution causes the vehicle to continue along an unstable S-shaped motion and collide with the vehicle from the opposite lane.
In Figure~\ref{fig:case_study_latency}(c), the 500ms delay further amplifies this temporal mismatch: the S-shaped trajectory becomes more pronounced and the ego vehicle collides with the vehicle on the right.
This case highlights that inference latency does not necessarily make the model predict obviously invalid waypoints; instead, it can make otherwise plausible actions arrive too late for the current scene state.
It complements the quantitative results in Table~\ref{tab:robust_full_combined}, where SimLingo shows strong clean performance but suffers severe degradation under delayed control execution.
\clearpage
\clearpage
\subsection{GPS Noise}
\begin{figure}[t]
    \centering
    \includegraphics[width=1\linewidth]{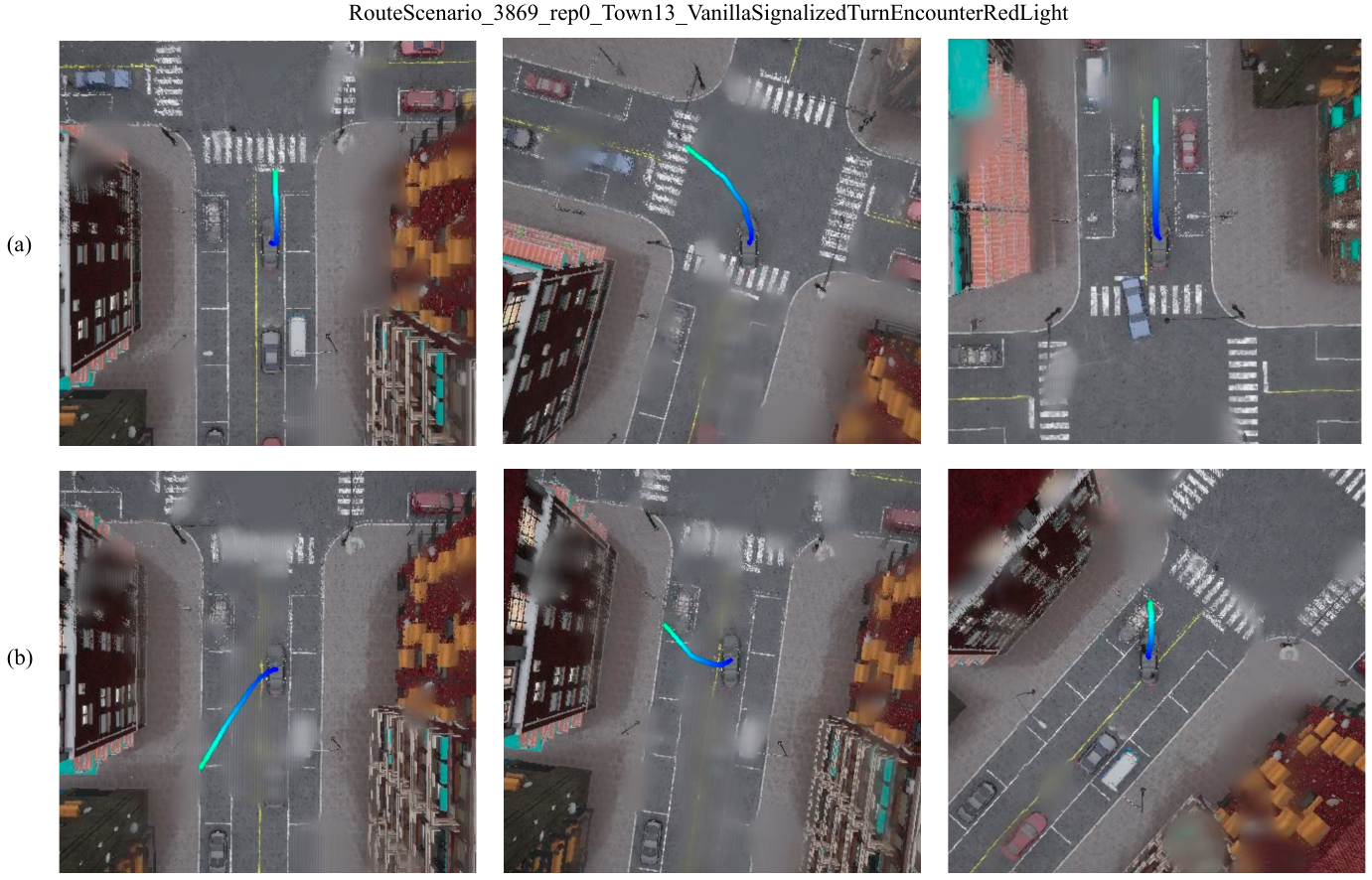}
    \caption{
    \textbf{Qualitative GPS localization-noise case study on TCP-traj.}
    We compare the same RouteScenario\_3869\_rep0 \texttt{VanillaSignalizedTurnEncounterRedLight} route under two settings:
    (a) clean baseline and
    (b) severe GPS localization noise with $\sigma_{\mathrm{GPS}}=15$m.
    The route requires the ego vehicle to approach a signalized intersection, respect the red-light constraint, and execute the intended turn with accurate route alignment.
    The overlaid BEV trajectory indicates the TCP-predicted waypoint trajectory used by the controller.
    Under severe GPS noise, the inferred ego-state and navigation target become less consistent with the true vehicle state, leading to degraded closed-loop behavior.
    }
    \label{fig:case_study_gps_tcp}
\end{figure}

Figure~\ref{fig:case_study_gps_tcp} shows a qualitative GPS localization-noise case study for TCP-traj on RouteScenario\_3869\_rep0, which belongs to the \texttt{VanillaSignalizedTurnEncounterRedLight} scenario in Bench2Drive.
In this route, the ego vehicle approaches a signalized intersection and must maintain route alignment while executing the intended turn under the traffic-light constraint.
Although the scenario is less visually complex than dense actor-flow cases, it requires accurate ego localization because the route planner and local target point are computed from the GPS-derived ego position.
Therefore, perturbing GPS can shift the inferred navigation target and cause the trajectory-following controller to execute a maneuver that is misaligned with the actual vehicle state.

Under the clean setting, TCP-traj receives a consistent ego-state/navigation signal and follows the intended turn.
With GPS localization noise, the predicted waypoint trajectory becomes misaligned with the true road geometry and route progress, causing the vehicle to deviate from the intended path and become blocked before completing the intersection maneuver.
This case complements Table~\ref{tab:robust_full_combined}, where severe GPS localization noise causes a large Driving Score and Success Rate drop for TCP-traj, illustrating that ego-state perturbations can induce concrete closed-loop failures even when camera observations remain available.

\clearpage
\clearpage
\subsection{Speed Noise}

\begin{figure}[t]
    \centering
    \includegraphics[width=1\linewidth]{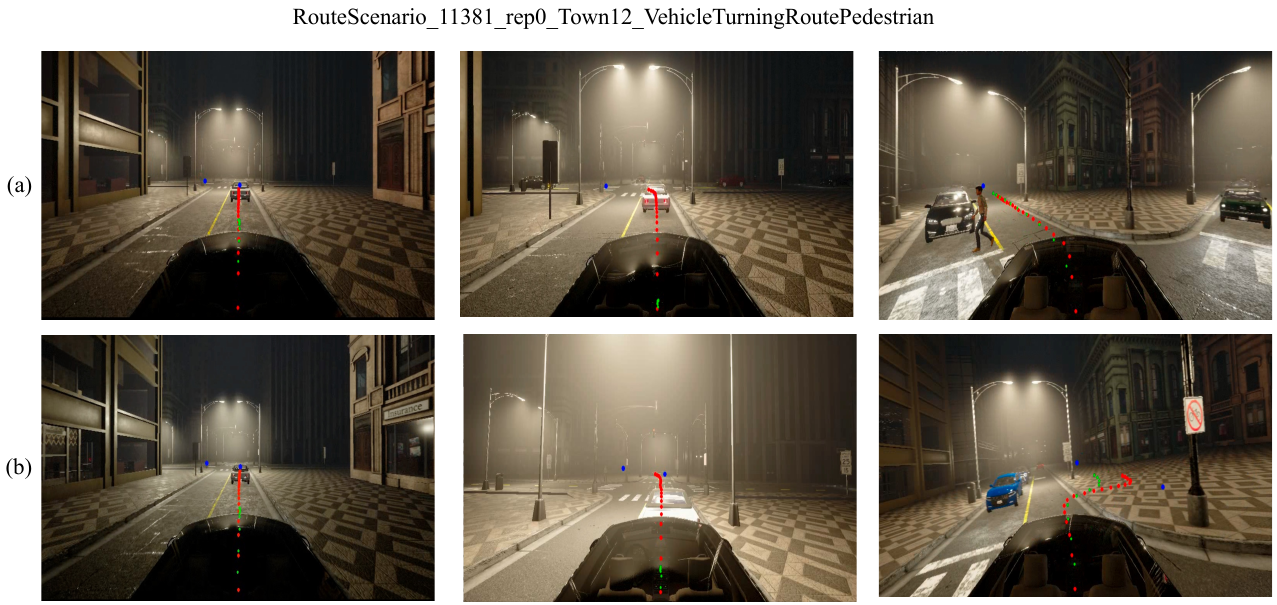}
    \caption{
    \textbf{Qualitative speed-noise case study on SimLingo.}
    We compare SimLingo on RouteScenario\_11381\_rep0, a \texttt{VehicleTurningRoutePedestrian} route, under:
    (a) clean baseline and
    (b) multiplicative speed noise with $\eta\sim\mathcal{N}(0.2,0.2^2)$.
    In the clean setting, the ego vehicle stops behind the leading vehicle at the red light, proceeds after the light turns green, and completes the left turn while yielding to the pedestrian.
    Under severe speed underestimation, the agent fails to brake sufficiently while waiting at the red light and later shows unstable waypoint predictions and aggressive vehicle body motion during the turning maneuver.
    \textcolor{blue}{Blue points} denote navigation target points,
    \textcolor{red}{red points} denote predicted path waypoints,
    and \textcolor{green!60!black}{green points} denote predicted speed waypoints.
    }
    \label{fig:case_study_speed}
\end{figure}

Figure~\ref{fig:case_study_speed} shows a qualitative speed-noise case study for SimLingo on RouteScenario\_11381\_rep0, which belongs to the \texttt{VehicleTurningRoutePedestrian} scenario.
This route requires the ego vehicle to approach a signalized intersection, stop behind the leading vehicle during the red phase, restart when the light turns green, complete a left-turn maneuver, and yield to the pedestrian.
It therefore depends on both semantic scene understanding and accurate longitudinal state estimation: the agent must decide not only where to go, but also how strongly to brake or accelerate at each stage.

In the clean baseline shown in Figure~\ref{fig:case_study_speed}(a), SimLingo handles the full interaction sequence correctly.
The ego vehicle stops behind the front vehicle at the red light, resumes motion after the light changes, and completes the left turn while yielding to the pedestrian.
In Figure~\ref{fig:case_study_speed}(b), severe multiplicative speed noise with $\eta\sim\mathcal{N}(0.2,0.2^2)$ causes the speed input to substantially underestimate the true vehicle speed.
As a result, the agent fails to judge the required braking strength while waiting at the red light and does not stop reliably behind the leading vehicle.
After the turn begins, the predicted waypoints also become less stable, and the rollout exhibits strong body motion during the turning phase.
This example complements Table~\ref{tab:robust_full_combined} by showing that speed noise can induce concrete longitudinal-control failures even when the visual scene remains available.

\end{document}